\documentclass[conference,letterpaper,10pt]{IEEEtran}

\usepackage{amsfonts}
\usepackage[colorlinks=true,citecolor=blue,linkcolor=blue,urlcolor=blue]{hyperref}
\usepackage{todonotes}
\usepackage{textcomp}
\usepackage{graphicx}
\usepackage{tikz}
\usepackage{enumitem}
\usepackage{wrapfig}
\usepackage{amsmath}
\usepackage[acronym]{glossaries}
\glsdisablehyper
\newacronym{bgm}{BGM}{Basal Ganglia Model}

\usetikzlibrary{arrows.meta, positioning, shapes.geometric}
\usepackage{booktabs}
\usepackage[capitalise]{cleveref}
\usepackage{multirow}
\usepackage{wrapfig}

\usepackage[ruled,vlined]{algorithm2e}
\usepackage{makecell}
\usepackage{soul}

\title{Bandit Algorithms for Deep Brain Stimulation}

\author{Arkaprava Gupta\IEEEauthorrefmark{1} \quad Nicholas Carter\IEEEauthorrefmark{1} \quad William Zellers\IEEEauthorrefmark{1} \quad Prateek Ganguli\IEEEauthorrefmark{1} \quad Benedikt Dietrich\IEEEauthorrefmark{2} \\ \quad Vibhor Krishna\IEEEauthorrefmark{3} \quad Parasara Sridhar Duggirala\IEEEauthorrefmark{1} \quad Samarjit Chakraborty\IEEEauthorrefmark{1}\IEEEauthorrefmark{4} \\
\IEEEauthorrefmark{1}Department of Computer Science, UNC Chapel Hill, USA \qquad
\IEEEauthorrefmark{2}Hochschule M\"unchen, Germany \\
\IEEEauthorrefmark{3}Department of Neurosurgery, UNC Chapel Hill, USA \qquad
\IEEEauthorrefmark{4}TU Munich Institute for Advanced Study, Germany
}

\begin{document}

\maketitle

\begin{abstract}
Deep Brain Stimulation (DBS) is an effective treatment for Parkinson’s disease, but conventional fixed-parameter stimulation can reduce battery life and cause side effects while failing to adapt to changing neural dynamics. Recent reinforcement learning approaches improve adaptability, yet most rely on deep neural networks that require offline training and are computationally too expensive for implantable hardware. This paper presents a resource-conscious adaptive DBS framework based on a Time- and Threshold-Triggered Pruned Multi-Armed Bandit (T3P MAB) algorithm. The proposed method jointly tunes stimulation frequency and amplitude, avoids prior training, and remains transparent enough to support clinician-guided adjustment. Using a computational basal ganglia-thalamic model, we show that T3P converges faster than competing MAB methods and outperforms deep-RL baselines in suppressing pathological beta-band activity while reducing stimulation power. We implemented it on different microcontrollers and report detailed energy measurements, showing convergence in under two minutes and suitability for resource-constrained implantable systems. These results support lightweight bandit-based control as a practical path toward personalized, energy-efficient DBS.
\vspace{1mm}

\end{abstract}


\section{Introduction}

\begin{wrapfigure}{r}{0.22\textwidth}
    \centering
    \includegraphics[width=0.22\textwidth]{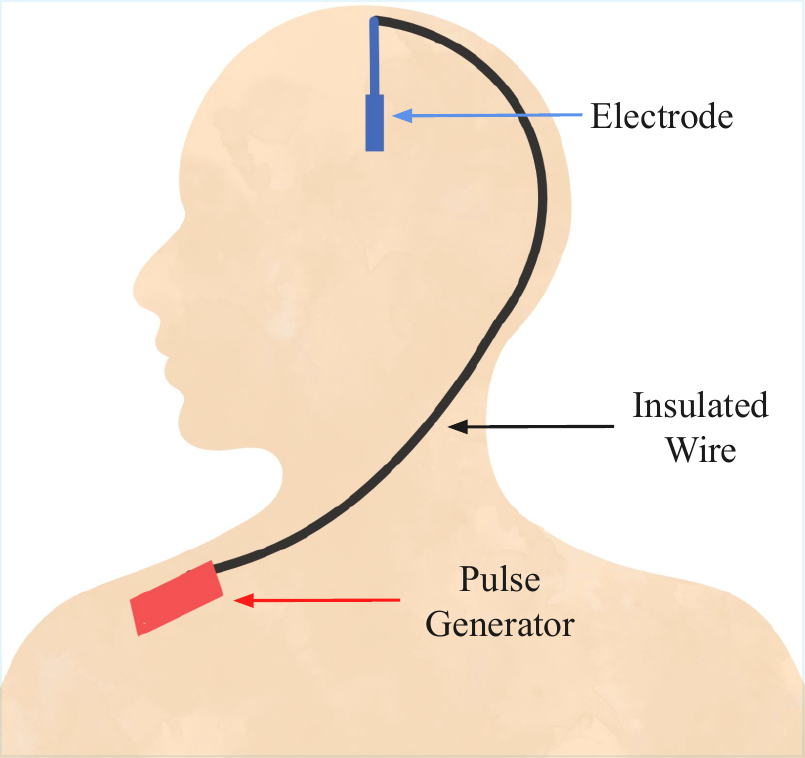}
    \caption{Deep Brain Stimulation}
    \label{fig:DBS}
\end{wrapfigure}

Around 930,000 people in the U.S. currently suffer from Parkinson's Disease  (PD) with 1,238,000 being the projected number by the year 2030 \cite{Marras2018}.
Patients with PD commonly  suffer from motor symptoms like tremors, rigidity, slowness of movement and postural instability. Deep brain stimulation (DBS) has proven to be a promising technique for the treatment of PD and is widely used today to alleviate motor symptoms in PD patients. Approximately 10,000 DBS implant surgeries are performed in the U.S. every year.
This procedure involves stimulating specific sites within the Basal Ganglia (BG) with electrical signals through an implantable device. This implant consists of a pulse generator and an electrode. The pulse generator is a battery-powered device responsible for generating the train of electrical pulses and is usually implanted under the skin near the collarbone. These pulses are delivered to the brain through the electrode, connected to the pulse generator by means of an insulated wire. The electrode is carefully placed inside the brain, with the tip at the specific site of the BG region. \cref{fig:DBS} provides an overview of this setup. This tightly coupled interaction between sensing neural activity, computation, and actuation through electrical stimulation forms a cyber-physical system (CPS), where embedded computation continuously interfaces with the physical dynamics of the human brain.

\vspace{0.5em}
\noindent 
\textbf{Types of DBS:}
Most modern clinical approaches use continuous DBS (cDBS) that involves stimulating the brain of the patient with a fixed amplitude and frequency. Patients must visit their neurologist multiple times a year so that the frequency of stimulations can be tuned to best fit their health needs. The effectiveness of stimulation is typically evaluated by monitoring neural activity in the brain regions affected by Parkinson’s disease, particularly those involved in motor control. In conventional setups, the strength of beta-band activity ($P_\beta$) is measured from local field potentials (LFPs) recorded in the globus pallidus internus (GPi) region of the brain. LFPs are recordings of the collective electrical activity of neurons that can be sensed via the implant. $P_\beta$ serves as a biomarker to guide the adjustment of DBS frequency in clinical setups. Details on the $P_\beta$ biomarker are given later in \cref{sec:preliminaries}.
 This approach, however, suffers from energy inefficiency \cite{Lattimore_Szepesvari_2020} and may cause unwanted side effects such as speech slurring. This has led to many works that focus on adaptive DBS (aDBS), which adjusts the stimulation based on the current state of the brain. aDBS is known to reduce side effects from stimulation while being energy efficient \cite{PI_controller_1, PI_controller_2}. Some commercial devices for aDBS exist, with most of them using either rule-based or SVM-driven control policies to adjust stimulation parameters.

\vspace{0.5em}
\noindent 
\textbf{aDBS Limitations:} Existing aDBS approaches have several limitations. Most approaches rely on deep-RL agents, which require an extensive offline training phase because they are too resource-intensive to run directly on the implant, as they rely on training neural networks. In offline training, patient data is collected from the implant and is used to train the deep-RL algorithms in simulations. The neural networks used in deep-RL algorithms also make the underlying decision process opaque, reducing interpretability and trustworthiness in clinical settings. Furthermore, these algorithms are computationally expensive, leading to higher energy consumption and shortened battery life. Another drawback is their lack of flexibility. Deep-RL policies are difficult to modify so that feedback from the clinician and/or the patient is included in the decision making algorithm. As PD progresses, the underlying neural patterns change, causing previous training data to become outdated and necessitating retraining of the RL agent. Currently, commercially available aDBS devices also need a calibration phase to personalize the stimulation parameter for each patient, as is required for deep-RL approaches, further increasing the burden on both clinicians and patients.

\vspace{0.4em}
\noindent 
\textbf{Our proposal:}
In order to alleviate these problems, we present a Multi-Armed Bandit (MAB) approach  denoted as the \textbf{Time \& Threshold-Triggered Pruned Multi-Armed Bandit (T3P MAB)} algorithm for DBS that is transparent and easier for the neurologist to understand and modify. It is efficient in terms of energy consumption and can be implemented \textit{in vivo} without the need for a prior offline training phase. MABs are a class of RL algorithms that do not model state transitions. They find the optimal action using observed rewards in the current state without the need for neural networks and are therefore lightweight in terms of computation. In our experiments, the T3P MAB agent demonstrates better performance with respect to both existing MAB and deep-RL approaches in suppressing PD symptoms. Notably, our MAB approach also uses less power for stimulation during this compared to deep-RL strategies. This is in addition to the advantage of sample efficiency, in contrast to deep-RL algorithms which require a huge amount of training data.

\vspace{0.5em}
\noindent 
\textbf{\textmu C implementations:}
We deployed our T3P MAB algorithm on the ESP32-S3 and ESP32-P4 microcontroller units (MCUs) and report power and energy consumption measurements using a high-precision real-time power analyzer. Our implementation is further compared against existing commercial DBS devices in terms of power consumption characteristics.
To our knowledge, this is the first work that deploys an MAB algorithm for DBS in hardware to report power metrics.
Our work advances the state-of-the-art in personalized medicine by minimizing the need for human intervention in regulating stimulation parameters while also ensuring symptoms suppression, with enough power-efficiency to be suitable for implantable hardware. Although MAB algorithms require a convergence time before the patient is stabilized, this duration is small, typically under $2$ minutes as observed in our experiments.

\vspace{0.5em}
\noindent 
\textbf{Our contributions}
are summarized below: 
\begin{enumerate}
    \item We propose our \textbf{T3P MAB} algorithm to explore the possibility of modulating amplitude and frequency for DBS, an approach which has not been explored extensively in literature.
    \item We compare the T3P MAB algorithm with other MAB approaches, as well as existing deep-RL approaches to show that our algorithm more effectively suppresses the $P_\beta$ biomarker and reduces DBS stimulation energy consumption more than the state-of-the-art deep-RL methods, with remarkable sample efficiency.
    \item We implement an MAB RL approach in hardware for the first time in the context of DBS and report detailed energy metrics. We further compare this implementation with both existing deep-RL methods and existing devices for DBS.
\end{enumerate}

\vspace{0.5em}
\noindent 
\textbf{Related work:}
Recent work has explored the possibility of using RL techniques to design effective strategies to perform aDBS with most of them relying on deep-RL approaches \cite{lu_2020, qitong2023, Cho, Gao2020, qitong2022, carter2025invivotrainingdeepbrain}. A majority of these only consider frequency modulation for DBS while the possibility of modulating amplitude, or both amplitude and frequency, has been left relatively unexplored \cite{Cho, Gao2023}.
Further, the use of MAB has not been sufficiently explored in the domain of DBS and only recently have some results on the use of MAB in restricted settings been reported~\cite{epsilon_neural_TS}. In contrast to these studies, where only the frequency of the stimulation has been adapted, we show that MAB can be successfully used for a more complete joint frequency and amplitude modulation.

A recent study compares various deep-RL algorithms used for DBS and finds the Twin Delayed Deep Deterministic Policy Gradient (TD3) algorithm to be superior to the others in terms of suppressing the $P_\beta$ biomarker \cite{Cho}. We used an implementation of the TD3 agent from a recent work \cite{carter2025invivotrainingdeepbrain} to draw a comparative study against our MAB approach. We also compare our algorithm against other MAB algorithms, along with the one proposed in \cite{epsilon_neural_TS} for frequency modulation only. The experiments are performed on a model of the human brain, that simulates both healthy and PD conditions, which has been refined through multiple works over the years \cite{Rubin,Pirini2009, relative_contributions, pajic2018}. Although RL has not been adopted into commercial devices yet, a recent work performs experiments on real patients to prove that the use of RL is indeed a promising direction of work \cite{Gao2023}.

\vspace{0.5em}
\noindent 
\textbf{Organization of the paper:}
In \cref{sec:preliminaries}, we introduce the details of the BGT model and discuss a variety of popular MAB algorithms. \cref{sec:methodology} proposes the problem formulation for MAB and the design of the reward function. Results are discussed in \cref{sec:experimental_results}, where we compare the MAB algorithms with existing deep RL algorithms in terms of effectiveness and energy efficiency. We conclude the paper and discuss future work in \cref{sec:conclusion}.

\section{Preliminaries}\label{sec:preliminaries}

In this section, we first describe the Basal Ganglia-Thalamic (BGT) model. The biomarkers used for testing effectiveness of DBS on the BGT model are discussed thereafter. Next, we discuss MAB as an RL algorithm and some of the popular MAB approaches that we use in our experiments. Finally, we discuss deep RL strategies and elaborate on the TD3 approach that has been frequently used in recent DBS research for PD.

\subsection{The Basal Ganglia-Thalamic (BGT) Model}

\begin{figure}[t]
    \centering
    \includegraphics[width=0.49\textwidth]{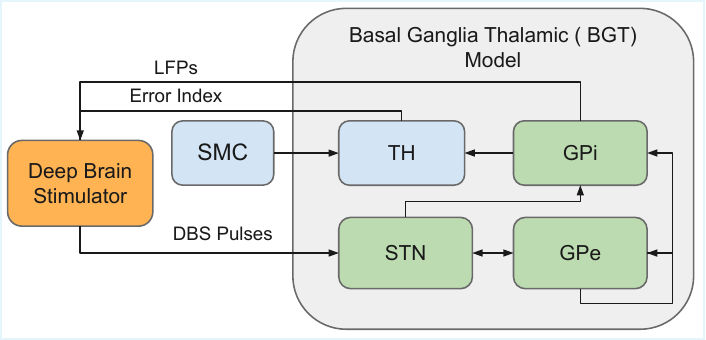}
    \vspace*{-0.7cm}
    \caption{The Basal Ganglia-Thalamic Model}
    \label{fig:BGT_model}
    \vspace*{-0.8cm}
\end{figure}

Due to the inherent complexities of testing black box neural network algorithms tasked with critical decision making on live human brains, we utilize a computational brain model \cite{Rubin}. More specifically, it models the BG and thalamus regions of the brain, as they are the most affected by PD. Additionally, the model is capable of simulating PD afflicted brain activity in addition to baseline neural activity. Several works have improved this model over the years \cite{Pirini2009, relative_contributions}, and we use the most recent one given in \cite{pajic2018}. The BGT model allows for simulation of voltage spiking activity of individual neurons of the globus pallidus internus (GPi), the globus pallidus externus (GPe), and the subthalamic nucleus (STN) regions found within the BG, as well as the thalamus (TH) region. A block diagram representation of the BGT model along with our setup is shown in \cref{fig:BGT_model}. 

The regions in green indicate regions of the BG that are part of our brain model, and the other regions in blue are ones that are not part of the BG but we include them in our model for a more realistic simulation of the human brain. As mentioned before, local field potentials (LFPs) can be obtained from the GPi region of the brain for calculation of the $P_\beta$ biomarker, which allows us to observe the current state of the brain. The $P_\beta$ is an important biomarker which is used by most of the state-of-the-art implementations of DBS algorithms today, not only in research but also in implants in the real patients. Similarly, the error index (EI) biomarker can be obtained from the TH region. We discuss both of these biomarkers in more detail at the end of this section. The deep brain stimulator is the device that resides in the human body which is responsible for sending impulses to the STN region of the brain to stabilize the patient. Although stimulating and sensing at different regions is not yet common in clinical practice, our setup is inspired by recent works that predominantly use this approach \cite{Cho, Gao2020, epsilon_neural_TS}. This configuration can further provide insights into performing DBS in the STN while sensing in the GPi using two different electrodes. Dealing with noise from sensing is a major problem in this domain. By separating sensing and stimulation, we might be able to offer more stable biomarkers for sensing and, therefore, deliver more effective stimulation. In our setup, an RL algorithm resides in this component that senses the current state of the brain using the aforementioned biomarker(s) and updates the stimulation parameters (frequency and amplitude in our case) adaptively.

The neural activity of each region is primarily determined by the following four equations (the change in membrane voltages of the $GPe$ and $GPi$ have been combined into a single equation): \textbf{(1)}~$C_m \frac{dv_{STN}}{dt} = -I_L - I_{Na} - I_K - I_T - I_{Ca} - I_{AHP}  - I_{GPe} + I_{app} + I_{DBS}(t)$, 
\textbf{(2)}~$C_m \frac{dv_{GPe/i}}{dt} = -I_L - I_{Na} - I_K - I_T - I_{Ca} - I_{AHP} 
     - I_{STN} + I_{GPe} + I_{app}$, and \textbf{(3)}~$C_m \frac{dv_{TH}}{dt} = -I_L - I_{Na} - I_K - I_T - I_{GPi} + I_{SM}(t)$.

Here, $C_m$ denotes the specific membrane capacitance of any given neuron. $dv_R/dt$ represents the change in membrane voltage over time in a region $R$, and when multiplied with $C_m$, represents the current flowing through a neuronal membrane in region $R$. $I_L$, $I_{Na}$, $I_K$, $I_T$, $I_{Ca}$, and $I_{AHP}$ denote the leak, sodium, potassium, low-threshold calcium, high-threshold calcium, and after hypo-polarization potassium channel currents, respectively. Variables of the form $I_R$, where $R \in \{GPi, GPe, STN, TH\}$, represent current flowing in from neurons in region R. 

$I_{app}$ 
represents the external currents influential to the BG regions. Decreasing the contribution of $I_{app}$ correlates with an insufficient dopamine supply from the substantia nigra pars compacta to the globus pallidus regions. This, in turn, leads to an increase in PD severity, as PD-afflicted neurons typically receive insufficient amounts of dopamine. $I_{SM}(t)$ denotes the current contributed by the sensorimotor cortex (SMC) region to the TH. SMC contribution is modeled as an array of anodal monophasic current pulses of amplitude $3.5$ $\mu A/\text{cm}^2$ and pulse width of $5$ ms. The frequency at which these pulses occur is randomized via selecting from a gamma distribution with an average of $14$ Hz and a coefficient of variation $=0.2$. 

Electrical stimulations are provided by our MAB algorithm to the BGT simulation via the $I_{DBS}$ variable. We stimulate the STN region of the BG, a common target of most modern day DBS implementations. Applying these electrical impulses to the STN is known to lower symptom severity in PD patients and correct neural misfirings \cite{striatal_networks}.

Now, we describe the Quality of Control (QoC) metrics which we consider for measuring the effect of DBS on the brain.
    \textbf{(1)~Error Index (EI):} The EI captures the amount of erroneous activations that occur in the TH region of the brain with respect to the input pulses from the SMC. In a healthy brain, impulses from the SMC activate all the neurons in the TH exactly once in an interval of 25 ms. This is, however, not the case in a PD brain. Unfortunately, the EI is not possible to obtain in real life with the current devices in use for DBS.
    \textbf{(2)~Beta-band Power Spectral Density ($P_\beta$):}  The GPi region of a PD brain exhibits pathological oscillations in the range of $13-35$ Hz band (known as the beta-band) which is less prevalent in a healthy brain. $P_\beta$ for a single neuron is given by $    P_\beta^j\left(x\right) =\int_{2 \pi \cdot 13 Hz}^{2 \pi \cdot 35 Hz} \left|\hat{f}\left(x_i\right)\right|\:df,
    $ The $P_{\beta}$ of the entire region of the GPi neurons is calculated as $ P_\beta = \frac{1}{n}\sum^n_{j=1} P_\beta^j
    $. The $P_\beta$ value can be obtained in the real world from LFPs and is strongly correlated with the EI, therefore we use the $P_\beta$ as feedback in our approach. A lower $P_{\beta}$ reading is correlated with decreased PD symptoms in the patient.

\vspace*{-.2cm}
\subsection{Reinforcement Learning Algorithms for DBS}
\vspace*{-.1cm}

\begin{wrapfigure}{br}{0.22\textwidth}
    \centering
    \hspace*{-0.3cm}\includegraphics[width=0.25\textwidth]{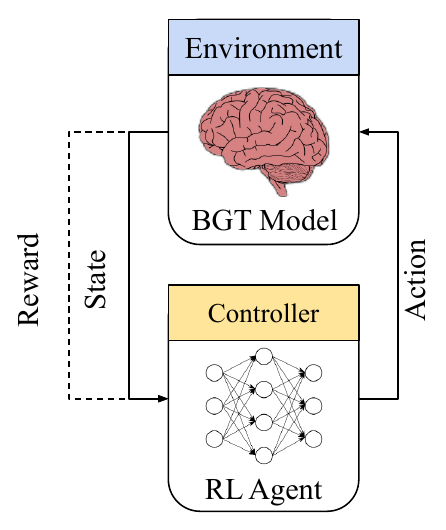}
    \vspace*{-0.8cm}
    \caption{\footnotesize{RL-BGT interaction.}}
    \label{fig:RL_BGT_interaction}
    \vspace*{-0.1cm}
\end{wrapfigure}

A variety of RL algorithms have been explored for DBS devices in literature, with deep-RL algorithms receiving more attention than the other RL approaches. Our work, however, focuses on the lightweight MAB algorithms for RL that make algorithms more suitable for DBS implants. First, this section elaborates on the deep-RL algorithms that have been widely studied in recent work and then describes the MAB algorithms we use in this paper.
In the context of DBS, the RL controller operates on the current neural state of the patient, represented as a vector of electrophysiological biomarkers. Formally, let the brain state at time step $t$ be denoted by $s_t = [b_1(t), b_2(t), \ldots, b_n(t)] \in \mathcal{S},$ where each $b_i(t)$ corresponds to a biomarker for DBS. The RL agent observes $s_t$ and selects a stimulation action $a_t = [f_t, A_t] \in \mathcal{A}$, where $f_t$ and $A_t$ represent the stimulation frequency and amplitude, respectively. The environment (i.e., the patient’s brain) responds to the stimulation with a new state $s_{t+1}$, and a scalar reward $r_t$ reflecting therapeutic efficacy. The objective of the RL controller is to learn a policy $\pi^*(a_t \mid s_t) = \arg\max_{\pi} \mathbb{E} \left[ \sum_{t=0}^{T} r_t \right],$ that maximizes the expected cumulative therapeutic reward while maintaining stability in the patient’s neural dynamics. This control objective differs from conventional RL applications, e.g., in games, where the agent aims to \textit{continuously manipulate} the environment to achieve a dynamically shifting goal state. Instead, in DBS, the aim is to \textit{identify and sustain} an optimal stimulation configuration $(f^*, A^*)$ such that the $P_\beta$ is minimized, i.e.,$(f^*, A^*) = \arg\min_{(f, A) \in \mathcal{A}} P_{\beta}(s_{t+1})$, thereby restoring and maintaining a stable, healthy brain state. The BGT model and the RL agent interaction is shown in \cref{fig:RL_BGT_interaction}. \\

\vspace*{-0.8em}
\noindent
\textbf{Deep Reinforcement Learning (Deep-RL):} Deep-RL algorithms combine reinforcement learning principles with deep neural networks to learn complex control policies directly from high-dimensional inputs. These algorithms aim to maximize cumulative reward by iteratively improving a policy that maps observations to actions, often through gradient-based optimization of a value or policy function. 

Among them, the Twin Delayed Deep Deterministic Policy Gradient (TD3) algorithm is a widely used actor–critic method designed to address overestimation bias in value learning. TD3 extends the Deep Deterministic Policy Gradient (DDPG) framework by employing two critic networks to compute a more reliable target value, delaying policy updates to stabilize learning, and adding noise to target actions for smoother value estimation. This enables TD3 to achieve robust and sample-efficient learning. \\

\vspace*{-0.8em}
\noindent
\textbf{Multi-Armed Bandit (MAB):} The Multi-Armed Bandit problem is a type of RL problem where a bandit learner must repeatedly choose between multiple actions with unknown rewards in an attempt to maximize cumulative rewards over time. 
Some of the popular MAB algorithms that we use in the context of our paper for choosing stimulation parameters for DBS are as follows. 

\noindent 
\textbf{(1)}~\emph{Upper Confidence Bound (UCB):} This algorithm attempts to select actions based not only on their estimated rewards but also on the uncertainty (confidence interval) associated with those estimates. The action to be taken is determined using the following equation:
        $a_t = \arg\max_a \left[ Q(a) + c \cdot \sqrt{\frac{\log t}{n_a}} \right]$, 
    where $t$ is the current timestep, $n_a$ is the number of times the arm $a$ has been played. The first term $Q(a)$ denotes the action-value estimate of arm $a$, a higher value which is indicative of an arm which has received higher rewards. The second term takes into account the uncertainty (specifically, the upper confidence bound), which decreases as the arm is played more frequently. The constant $c$ is a tuning parameter that scales the confidence interval. A larger value results in an increased likelihood of exploring actions with less certain rewards, whereas, a smaller value makes it inclined towards actions that seem to be more optimal based on the current knowledge. UCB naturally decays exploration over time.

\noindent 
\textbf{(2)}~\emph{Thompson Sampling (TS):} 
Thompson Sampling uses a \; Bayesian approach to action selection. 
It maintains a probability distribution over each action's expected reward, capturing both the estimated value and the uncertainty in that estimate. 
At each timestep, a possible reward is sampled from this posterior distribution for every arm, and the arm with the highest sampled value is chosen. 
This method naturally balances exploration and exploitation, since actions with high uncertainty may occasionally yield large sampled rewards. The action selection rule is given by
$a_t = \arg\max_a \tilde{\theta}_a, \text{where } \tilde{\theta}_a \sim \mathcal{N}(\mu_a, \sigma_a^2).$ Here, $\tilde{\theta}_a$ is a random sample drawn from the posterior distribution of the expected reward for arm $a$, modeled as a Gaussian with mean $\mu_a$ and variance $\sigma_a^2$. The parameters $\mu_a$ and $\sigma_a^2$ are updated based on the observed rewards and counts after each action. For more details, we direct the readers to~\cite{Lattimore_Szepesvari_2020}.

\noindent 
\textbf{(3)~$\epsilon$-Greedy:} The $\epsilon$-greedy algorithm is a simple yet effective heuristic that balances exploration and exploitation by introducing randomness into the decision process. The algorithm chooses with probability $\epsilon \in [0,1]$, an action uniformly at random (exploration), and with probability $(1-\epsilon)$, selects the action with the highest estimated mean reward (exploitation). At each timestep $t$, the agent selects:
    \begin{equation*}
        a_t =
\begin{cases}
\displaystyle \arg\max_a Q(a), & \text{with probability } 1 - \epsilon, \\
\text{random arm}, & \text{with probability } \epsilon.
\end{cases}
    \end{equation*}

Generally, the implementation of this algorithm involves starting with an initial $\epsilon$ value and decaying the value of $\epsilon$ by a certain factor until it reaches zero, after which only the arm with the highest estimated mean reward is used. This ensures that the algorithm converges to an optimal action. 

\noindent 
\textbf{(4)~Neural MAB:}
    The Neural MAB extends traditional bandits by using a neural network to approximate the reward function, allowing it to capture nonlinear relationships between context and reward. At each timestep $t$, the network estimates the expected reward $Q(a, x_t)$ for each arm $a$ and updates its parameters $\theta$ based on the observed reward $r_t$. The action is selected as $ a_t = \arg\max_a , f_\theta(x_t, a) + \eta_t,$ where $f_\theta(\cdot)$ is the predicted reward and $\eta_t$ is an exploration term. This approach enables generalization across arms and adapts effectively in complex or non-stationary environments, combining deep learning’s expressiveness with classical bandit exploration.

\section{Methodology}\label{sec:methodology}

We begin by presenting our formulation of the MAB problem for selecting appropriate stimulation parameters in DBS. Next, we outline the reward function employed in our implementation. We conclude with describing our algorithm.

\subsection{Problem Formulation}

The frequency and amplitude selection task is formalized as a sequential decision making problem under uncertainty, where the goal is to maximize therapeutic benefit while minimizing side effects. Specifically, we model each combination of candidate stimulation frequency $f_p \in \mathcal{F}$ and amplitude $a_q \in \mathcal{A}$ as an arm $(f_p,a_q)$ in a MAB setting, where $p \in \{1, 2, ..., m\}$ and $q \in \{1, 2, ..., n\}$ index the $m$ candidate frequencies and $n$ candidate amplitudes respectively. Therefore, we derive a set of candidate actions
$\mathcal{X} = \{ (f_i, a_j) \mid i = 1, \dots, m; \; j = 1, \dots, n \}, |\mathcal{X}| = m \cdot n $.
\noindent At every decision point $t$, an arm $(f_{p,t}, a_{p,t})$ is chosen and a reward is obtained based on the outcome which is then used by the MAB algorithm to learn how the arms compare in terms of effectiveness, which in our context is the overall positive outcome of the patient.

\subsection{Reward Function} \label{sec:reward_function}
We describe the formulation of our reward function here. The playing of every arm follows an instantaneous reward.
In our setup, the instantaneous reward after using an arm is given by $R_\text{instant}(t) = \alpha \cdot r_1 + \beta \cdot r_2 + \gamma \cdot r_3 \text{,}$ which is a function of the following components:
    \textbf{(1)~PD Intensity $(r_1)$:} As mentioned earlier, a lower $P_\beta$ is indicative of reduced PD symptoms. Thus, we calculate PD intensity from the $P_\beta$ reading during the application of the arm and incorporate it into the reward function with $r_1 = P_\beta$.
    \textbf{(2)~Continuous periods of no stimulation:} It has generally been considered safer to provide DBS with longer periods of no stimulation to the human brain, that is, with longer pulse intervals. The second component of our reward $r_2$ is calculated as 
    \begin{equation}
    r_2 =
    \begin{cases}
    1, & \text{if } I_\text{DBS}[t] = 0 \\
    0, & \text{otherwise}.
    \end{cases}
    \end{equation}

Where $I_\text{DBS}[t]$ is the amplitude of DBS current provided to the brain at time step $t$. Note that the duration of a single stimulation period in our implementation is $0.01$ ms. 
    \textbf{(3)~Energy Consumption:} Since the device runs on a battery, we would like to conserve the battery life of the device to minimize the need for frequent charging and therefore surgeries for battery replacements in the long run. Also, it is favorable for the health of the patient to provide only as much stimulation to the brain as necessary for their well-being. Therefore, we incorporate energy consumption into our reward, calculated as
    \begin{equation}
         r_3 = I_{\text{RMS}} = \sqrt{\frac{1}{T} \int_0^T I_{\text{DBS}}^2(t) \, dt}\textbf{,} \label{eqn:power_consumption}
    \end{equation}
    where $T$ is the total duration of $I_{DBS}$ of the STN region.
\noindent
Note that all the components of the reward function are scaled to the range $[0, 1]$ to ensure stability of the algorithm. The coefficients $\alpha$, $\beta$ and $\gamma$ used are $-0.7$, $0.1$ and $-0.2$, respectively.

\subsection{Proposed Approach: T3P Bandit} \label{sec:proposed_approach}
\vspace*{-0.1cm}
In our experiments, a few of the existing MAB algorithms have demonstrated the ability to converge to optimal stimulation parameters with the right choice of parameters in a reasonable time frame. However, as with many RL frameworks, such algorithms can be further improved by adapting to the unique characteristics of their target domains. Through our experiments, we observe that the mean reward obtained from each arm can be sufficiently estimated by playing all the arms just once. Furthermore, we find that arms among these which returned below-average rewards never turned out to be the optimal arm and were always ones that are unsuitable for DBS in the current state. We use this domain knowledge to add a pruning technique to the well-known $\epsilon$-greedy bandit algorithm to create our Time \& Threshold-Triggered Pruned (T3P) MAB algorithm. Our modification to the existing algorithm restricts its exploration to a set of arms which are good candidates for the stimulation parameters thus reducing instances of stimulation using non-optimal arms during exploration while also leading to faster convergence.

The working of our T3P MAB agent is given in Algorithm~1. It starts with a warm-up phase where it plays all arms once to get an estimate of the associated rewards. The $\epsilon$ parameter does not start decaying until the warm-up phase is complete. At the end of the warm-up phase, all arms except the top K are pruned, with the bandit only able to choose from the top K. 
At every round, an arm $i$ is played and the LFPs from the GPi region is recorded consequently. The $P_\beta$ is calculated from these recorded LFPs as given in \cref{sec:preliminaries} and it is used to compute the reward for that arm along with characteristics of the DBS current as discussed earlier in \cref{sec:reward_function}. This reward is used to update the value estimate $Q_i$ of the arm $i$. A higher $Q$ value indicates a more effective arm.
With the decaying of $\epsilon$, the algorithm converges to the optimal arm. Given that the T3P bandit has a high convergence rate, it is triggered again once a countdown timer elapses. This is done in order to update the DBS signal to adapt to the brain state in case it has changed but is not observable through the biomarker. The other scenario when it is triggered is when a change in the state of the brain (through the $P_\beta$ biomarker) is observed. This might be detected by a deviation from the reading of the $P_\beta$ biomarker by a certain threshold, in which case, the T3P algorithm will be triggered again in an attempt to find the new optimal arm. The parameter $K$ is tunable. Details on tuning this parameter is discussed later in \cref{subsec:hyperparameter_tuning}. Subsequently, we compare our proposed algorithm against other existing MAB approaches for DBS in \cref{subsec:compare_with_other_MABs}. 

The work most closely related to ours is a recent study that employs a modification of the neural TS MAB algorithm to tune only the frequency of stimulations \cite{epsilon_neural_TS}. With our best implementation of this method, it frequently converges to a suboptimal arm. In instances where it does reach the optimal arm, it uses non-optimal parameter settings for prolonged time periods, which could result in unfavorable patient experiences.
Next, we discuss the T3P MAB results, comparing them with other MAB and deep-RL approaches.

\section{Experimental Results}\label{sec:experimental_results}

Here, we discuss our implementation on hardware and report power measurements. We then compare these power measurements with existing approaches to show the effectiveness of our approach in terms of not only performance but also energy.

\begin{figure*}[t]
    \centering
    \begin{minipage}[t]{0.32\textwidth}
        \centering
        \includegraphics[width=\linewidth]{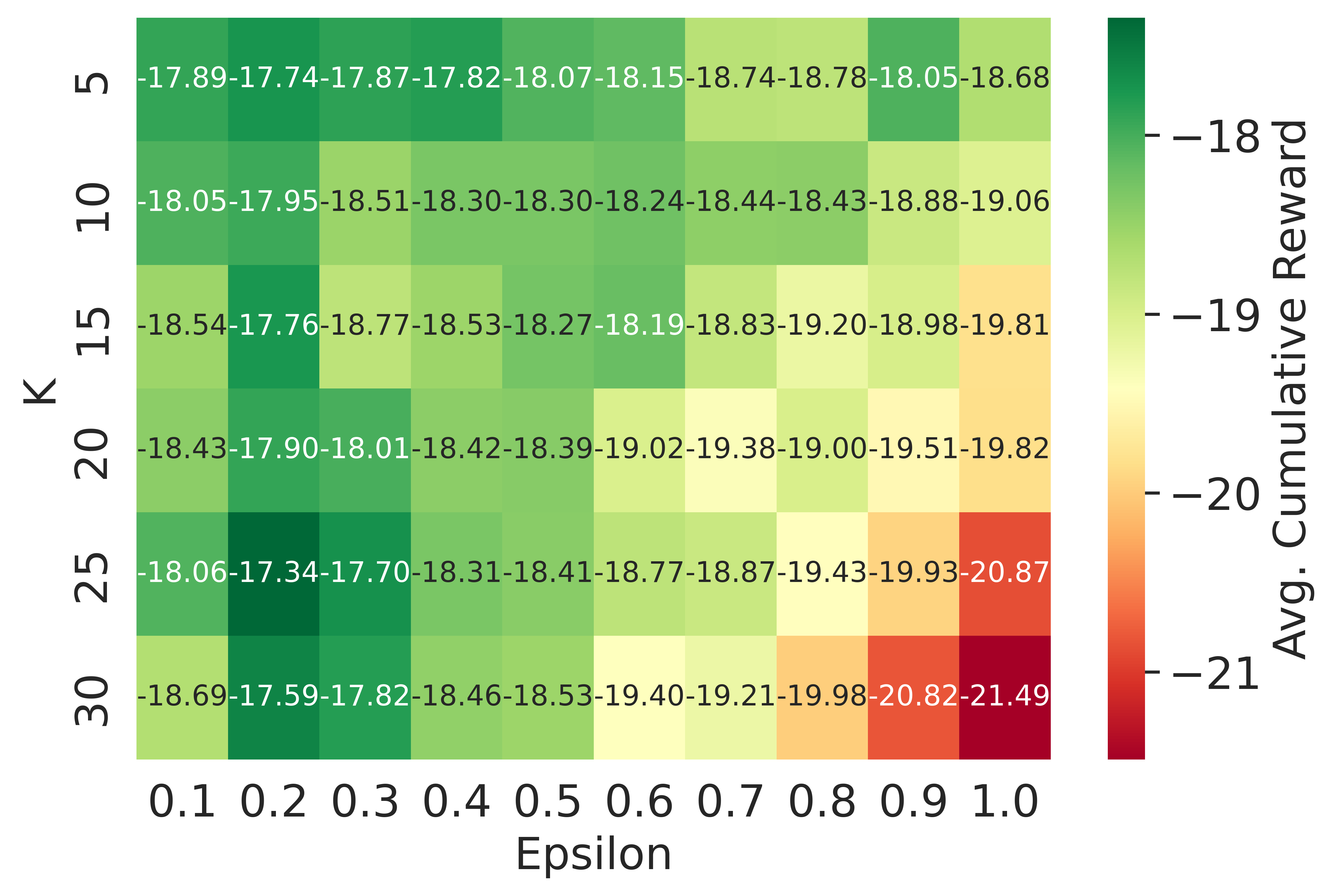}
        \vspace*{-.8cm}
        \caption{Hyper-parameter tuning of T3P MAB}
        \label{fig:heatmap}
    \end{minipage}%
    \hfill
    \begin{minipage}[t]{0.335\textwidth}
        \centering
        \includegraphics[width=\linewidth]{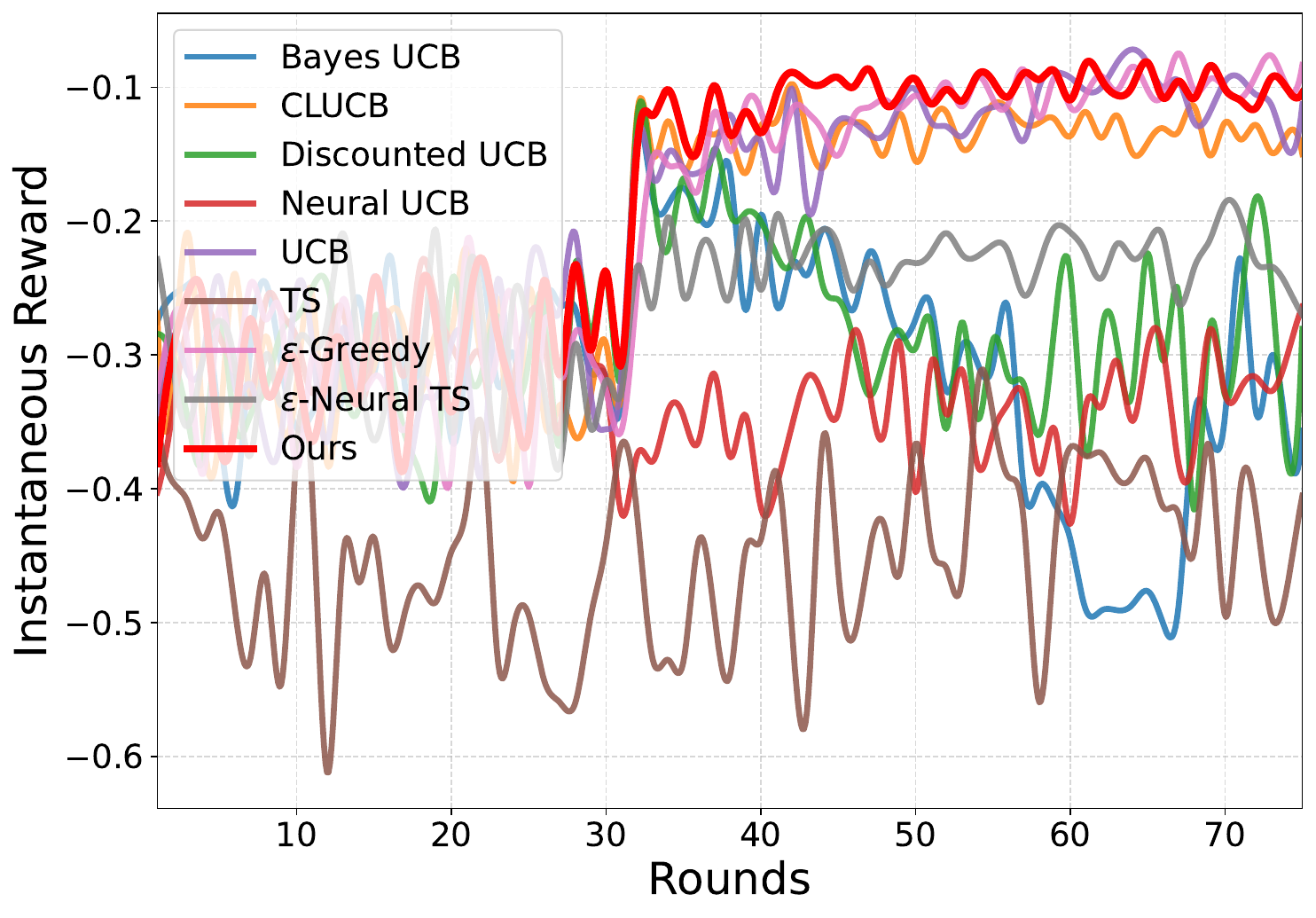}
        \vspace*{-.8cm}
        \caption{Comparison using rewards}
        \label{fig:MAB_comparison_instant_rewards}
    \end{minipage}%
    \hfill
    \begin{minipage}[t]{0.335\textwidth}
        \centering
        \includegraphics[width=\linewidth]{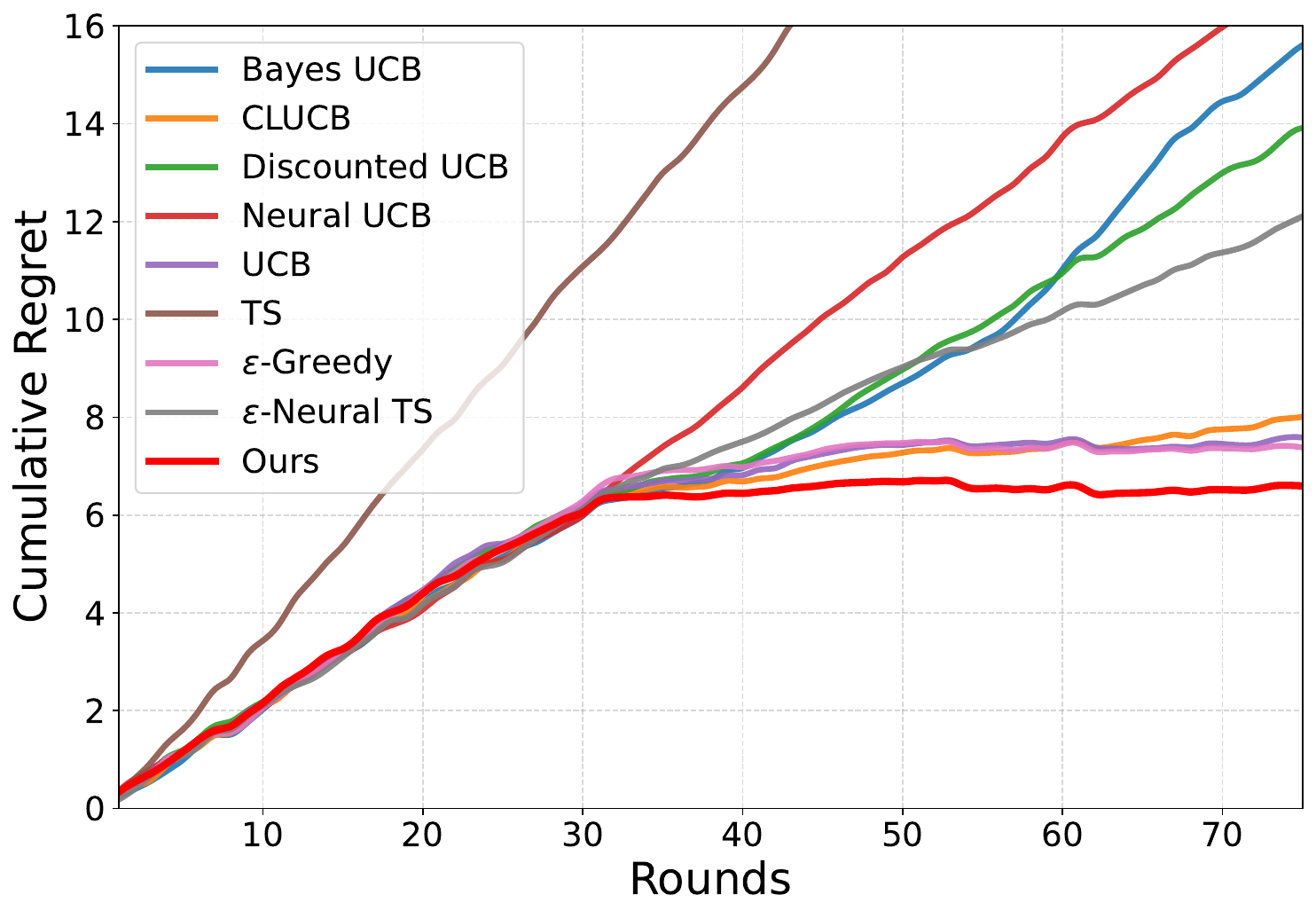}
        \vspace*{-.8cm}
        \caption{Comparison using cumulative regret}
        \label{fig:regret}
    \end{minipage}
    \vspace*{-0.5cm}
\end{figure*}

\subsection{Experimental Setup} \label{experimental_setup}
As mentioned earlier, we perform the experiments on the computational BGT model from \cite{pajic2018}. We use frequencies in the range $[55,180]$ Hz as frequencies lower than $50$ Hz have shown to be ineffective for DBS. Amplitudes that the algorithm could use were restricted to the range $[0,5000]$ $\mu A/ cm^2$. Works regularly implement DBS with these set ranges and have found them safe for \emph{in-vivo} use \cite{Cho,carter2025invivotrainingdeepbrain}. Our MAB algorithm attempts to choose from $31$ arms which are combinations of candidate frequency values $\{ 55, 80, 105, 130, 155, 180 \}$ Hz and candidate amplitude values $\{0, 1000, 2000, 3000, 4000, 5000 \}$ $\mu A/ cm^2$. The pulse duration used for DBS is $300$ $\mu s$. We use a bi-phasic current for charge balancing in the brain. Charge balancing is necessary as it lowers the risk of negative faradic reactions with brain tissue \cite{Piallat}. The DBS waveform is therefore symmetric where an anodic current for $150$ $\mu s$ is followed by a cathodic current for $150$ $\mu s$. The length of each round in our implementation is $1000$ ms.

With the setup described above, we analyze the performance of our algorithm on the following hardware platforms:

\begin{enumerate}\item The \textbf{ESP32-S3} uses a processor with dimensions $5$ mm x $5$ mm and is suitable for a device that needs to implanted inside the human body. The devices uses a dual-core Xtensa LX7 CPU with a maximum CPU frequency of 240 MHz, and contains $512$ kB SRAM and $384$ kB ROM. It further supports Wi-Fi and Bluetooth connectivity for communication with external devices to monitor health metrics of the patient. The system operates at a nominal voltage of $3.3$ V, consistent with standard low-power embedded platforms. 

\item The \textbf{ESP32-P4} which comes with a dual-core RISC-V CPU that runs up to $400$ MHz, $768$ kB SRAM and a low power core that can run at $40$ MHz. This MCU too comes in a size that can be used for implantable devices with the dimensions of the processor being $10$ mm x $10$ mm.
\end{enumerate}

To accurately measure the power consumption of these MCUs during operation, we use the \textbf{Joulescope JS220} Precision Energy Analyzer. The JS220 enables high-resolution, real-time measurement of various power consumption metrics, allowing precise computation of instantaneous power and total energy consumption over time. Data was recorded using the Joulescope software interface to capture detailed energy profiles of the ESP32 MCUs under various DBS algorithms.

\begin{algorithm}[t]
\caption{T3P MAB for DBS Parameters Selection}
{\scriptsize
\SetKwInOut{Input}{Input}\SetKwInOut{Output}{Output}
\Input{Number of arms $A$, decay parameters $(\epsilon_{start}, \epsilon_{min}, decay\_steps)$, top-$K$ arms $K$, biomarker $P_\beta$, deviation threshold $\delta$}
\Output{Optimal arm $a^*$ for stimulation}
\DontPrintSemicolon

\SetKwFunction{TTP}{T3P}
\SetKwProg{Fn}{procedure}{}{}

\Fn{\TTP{$A, \epsilon_{start}, \epsilon_{min}, decay\_steps, K, P_\beta, \delta$}}{

    \tcp{Warm-up Phase: Play all arms once}
    \For{$a = 1$ \KwTo $A$}{
        Play arm $a$, observe reward $r_a$\;
        Update estimated reward $Q(a)$ and count $n(a)$\;
    }
    
    \tcp{Prune arms: keep only top K based on rewards}
    Sort all arms by $Q(a)$ in descending order\;
    Keep only top $K$ arms, discard the rest\;
    
    \tcp{Start epsilon decay only after warm-up}
    $\epsilon \leftarrow \epsilon_{start}$\;
    $t \leftarrow 0$\;
    
    \While{True}{
        $t \leftarrow t + 1$\;
        
        \tcp{$\epsilon$-greedy selection among top K arms}
        Generate $u \sim U(0,1)$\;
        \eIf{$u < \epsilon$}{
            Select a random arm $a_t$ from top-$K$\;
        }{
            $a_t \leftarrow \arg\max_{a \in K} Q(a)$\;
        }
        
        Play arm $a_t$, observe reward $r_t$\;
        Update $Q(a_t)$ and $n(a_t)$ based on $r_t$\;
        
        \tcp{Decay epsilon}
        $\epsilon \leftarrow \max(\epsilon_{min},\; \epsilon_{start} - \delta \cdot t)$\;
    }   
    \Return{$a^* = \arg\max_{a \in K} Q(a)$}\;
}
}
\end{algorithm}

\subsection{Hyper-parameter Tuning of the T3P MAB algorithm}\label{subsec:hyperparameter_tuning}
We will now discuss how we tune the hyperparameters of the T3P MAB algorithm, namely the exploration rate ($\epsilon$) and the number of top arms K considered for DBS after the warmup phase. As discussed in section \cref{sec:proposed_approach}, the T3P MAB algorithm prunes arms which are less likely to be effective for DBS in order to reduce the search space of the MAB algorithm. As usual, $\epsilon$ is the exploration probability. We perform a grid search for finding the optimal combination of $\epsilon$ and K. It can be seen from \cref{fig:heatmap} that the highest cumulative average reward is obtained by a combination of $\epsilon = 0.2$ and $K=25$. The average cumulative rewards calculation is based on $10$ runs of the T3P MAB algorithm for each $\epsilon$ and K combination. A higher average reward value is marked with a greener value with lower values marked in red. We experiment with K values in the range $[5,30]$ as we have $31$ arms in total. A K value of $30$ means that only one arm is pruned at the end of the warmup stage and the algorithm can explore from a total of $30$ arms after that. The heatmap shows the trend of a lower $\epsilon$ being able to accumulate more rewards. This is because a higher exploration probability may result in playing more non-optimal arms unnecessarily. However, this trend is broken at $\epsilon=0.1$ because extremely low exploration can result in convergence to a non-optimal arm. The choice of top K restricts the exploration space. For higher exploration probabilities, a lower value of K can be seen to gather more rewards although this is less prominent for $\epsilon$ values under $0.4$. For our implementation, we linearly decay $\epsilon$ by $0.025$ after each episode.

\subsection{Comparison with other MAB approaches}\label{subsec:compare_with_other_MABs}
We compare our proposed T3P agent against $8$ other MAB algorithms, namely Upper Confidence Bound (UCB), Bayes UCB \cite{bayes_ucb}, Discounted UCB \cite{discounted_ucb}, Neural UCB \cite{neural_ucb}, CLUCB \cite{clucb}, Thompson Sampling (TS), $\epsilon$-Greedy and the $\epsilon$-Neural Thompson Sampling technique in \cite{epsilon_neural_TS}. Note that $\epsilon$-Neural Thompson sampling was proposed for a setting where only the frequency of the DBS current could be modulated. We modify its implementation for our experiments to allow it to change both frequency and amplitude of the stimulations. 
The results of these algorithms are based on their best hyper-parameter settings. We use $c = 0.05$ for UCB, $c=1.0$ for Bayes UCB, $c=0.6$ and $\alpha =0.05$ for CLUCB, $c=0.35$ and $\gamma = 0.99$ for Discounted UCB, $\alpha = 0.9$ and $\lambda =0.75$ for Neural UCB, $\epsilon = 0.4$ for $\epsilon$-Greedy, $\epsilon = 0.1$ for $\epsilon$-Neural TS.

\begin{figure}[t]
    \centering
    \includegraphics[width=0.5\textwidth]{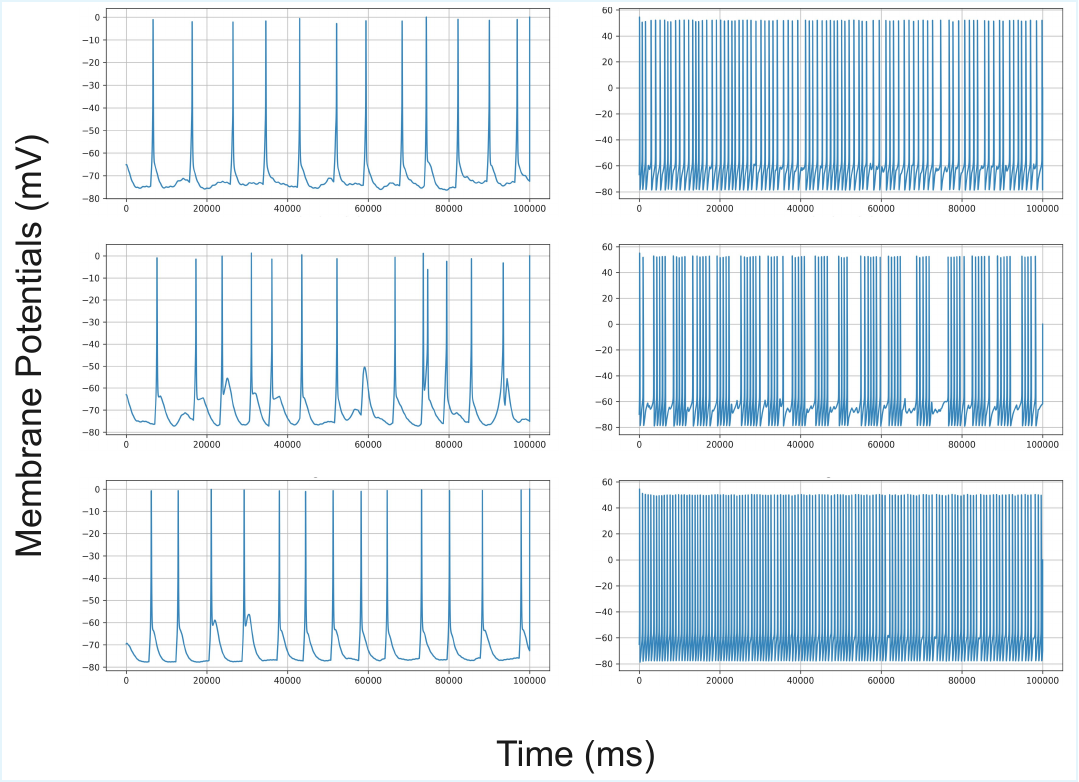}
    \vspace*{-.8cm}
    \caption{Neuron activity in the TH (left) and GPi (right) in a healthy brain (top row), a PD brain (middle row) and a PD brain with DBS (bottom row) }
    \label{fig:neuron_activities_with_dbs}
    \vspace{-5mm}
\end{figure}

The comparison of their performance in terms of average instantaneous rewards obtained over $75$ rounds is shown in \cref{fig:MAB_comparison_instant_rewards}. The values on the x-axis are the average instantaneous rewards based on $10$ runs for each algorithm. The instant reward is derived from the proposed reward function that also serves as feedback to the algorithms. As shown in the graph, the T3P MAB algorithm exhibits the fastest convergence toward the optimal arm and consistently outperforms other algorithms in terms of instantaneous rewards and convergence speed. $\epsilon$-Greedy, UCB and CLUCB, although worse than T3P, were close to each other with regard to the same metrics. The $\epsilon$-Neural TS often converged to a suboptimal arm. However, the other MAB algorithms failed to converge by the end of $75$ rounds.

The cumulative regret of each of these algorithms over time based on the average of $10$ runs of the algorithms is shown in \cref{fig:regret}. Instantaneous regret at time $t$ is calculated as the difference between the reward obtained by the optimal arm, if it was played in that round, and the reward obtained by the arm the MAB algorithm actually played at time $t$. The UCB, Epsilon-Greedy and our proposed T3P MAB algorithms all converge to the optimal arm, with the T3P algorithm converging more quickly than the others. The other MAB algorithms failed to converge to the optimal arm in almost all cases. The optimal arm chosen for regret calculation was the one that used a frequency of $155$ Hz and an amplitude of $1000$ $\mu A/ cm^2$ based on empirical observations.

\begin{figure*}[t]
    \centering
    \begin{minipage}[t]{0.33\textwidth}
        \centering
        \includegraphics[width=\linewidth]{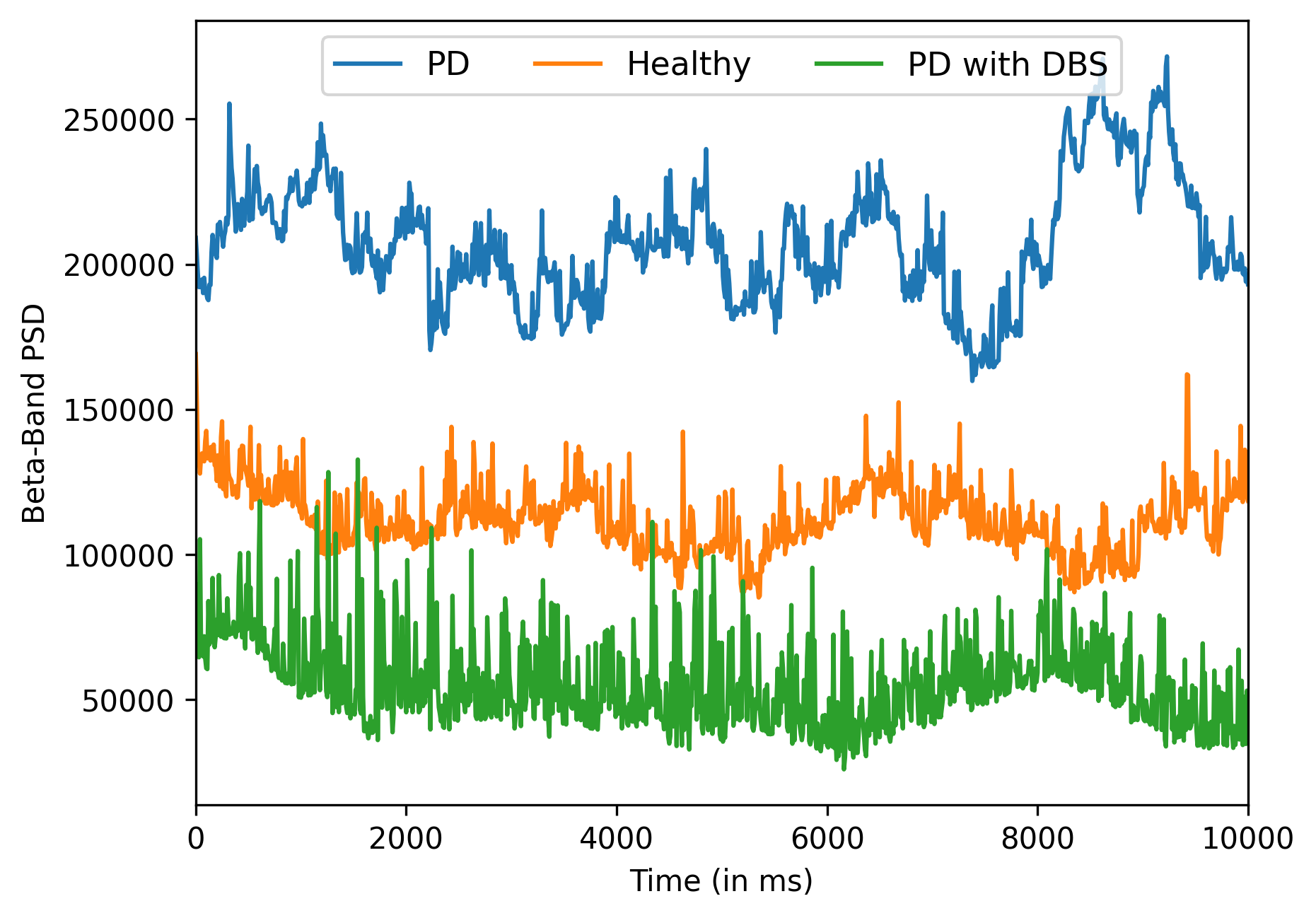}
        \vspace*{-0.8cm}
        \caption{$P_\beta$ suppression with DBS from T3P MAB.}
        \label{fig:signals_plot}
    \end{minipage}%
    \hfill
    \begin{minipage}[t]{0.66\textwidth}
        \centering
        \includegraphics[width=\linewidth]{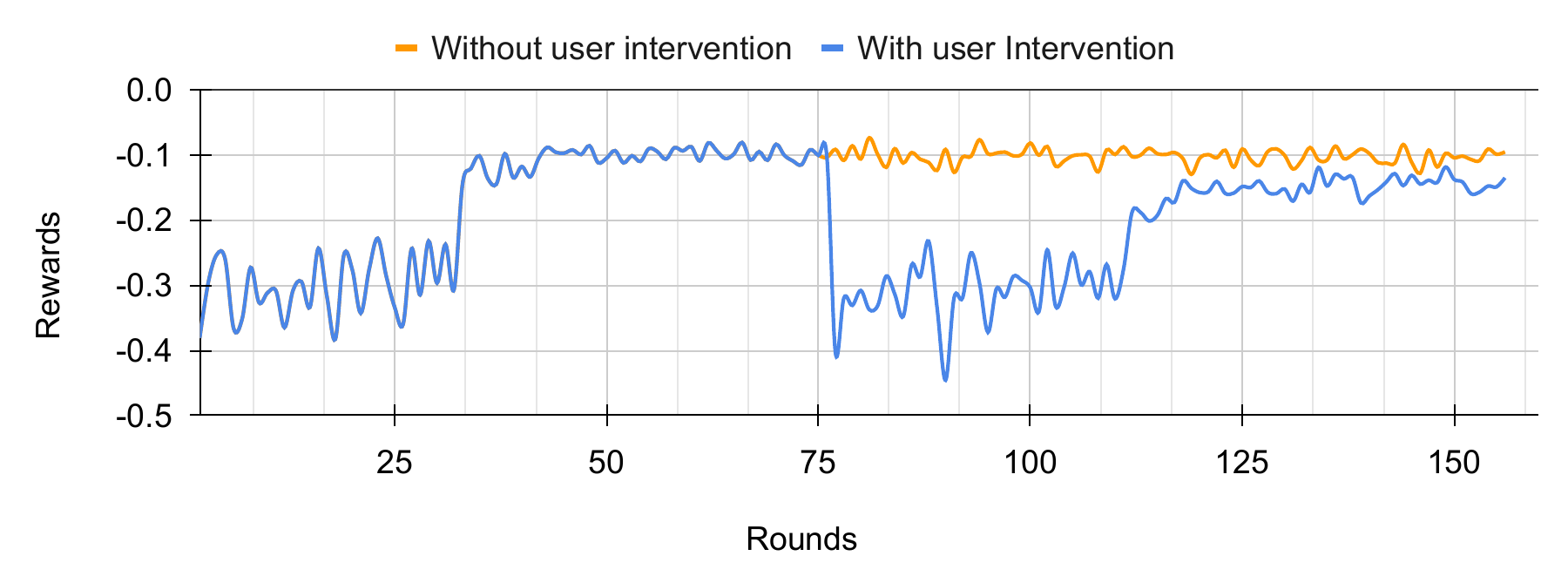}
        \vspace*{-0.8cm}
        \caption{Adaptability of T3P when patient intervention is introduced due to reported discomfort.}
        \label{fig:user_intervention}
    \end{minipage}
     \vspace*{-0.4cm}
\end{figure*}

We observe that $neural$ variations of MAB algorithms, in general, fail to perform as well as the other MAB algorithms. We believe this is due to the fact that it involves training a neural network to estimate the rewards from actions. Neural network components, being sample inefficient, require more rounds to converge. Also, the possibility of a change in state of the brain is not considered in \cite{epsilon_neural_TS} in which case, the neural network would have to be trained from scratch again. In \cref{fig:neuron_activities_with_dbs}, the activity of the membrane potentials in a single neuron is shown from our simulation of a healthy and a PD brain. The figure also shows how the activity in a PD brain changes when DBS is enabled with the T3P MAB agent. The characteristics of a PD brain, which are missed spikes in neurons in the TH region and occurrence of spikes in bursts in the GPi region are shown in the left and right figure in the middle row respectively. In the figures shown in the last row, both these characteristic features vanish with DBS applied to the brain. \cref{fig:signals_plot} shows the $P_\beta$ reduction with the T3P agent after it has converged. It can be noticed that the $P_\beta$ never crosses the value from that of a normal brain which is indicative of the fact that the patient symptoms are suppressed. 

\subsection{Comparison with existing deep-RL methods}

We use the implementation of the Twin-Delayed Deep Deterministic Policy Gradient (TD3) agent \cite{carter2025invivotrainingdeepbrain}, which is the state-of-the-art in DBS for PD among other deep RL algorithms and establish that the more computationally efficient MAB algorithms show performance comparable to deep RL methods in terms of suppressing the $P_\beta$ biomarker and energy used for generating DBS signals.

However, it should be noted that that the technique used for $P_\beta$ computation in their work is different than how we do it for the experiments. The $P_\beta$ calculation for the TD3 agent was done by computing the $P_\beta$s of individual neurons from membrane potential readings (accessible from the computational BGT model) and averaging them to find the $P_\beta$ which they report in their results. In our experiments, we compute the mean of the membrane potentials of the $10$ neurons to get the LFP which is then used to compute $P_\beta$. The technique we use can be considered more realistic as the former technique requires precise sensors (in order to extract membrane potential readings from singular neurons) which are more expensive and uncommon for DBS implants. LFPs, which are recordings of membrane potentials from a group of neurons, are simpler to obtain and commonly used in practice so we use this method to compute the $P_\beta$ biomarker. In order to compare between the techniques, however, we resort to the former method for computing the biomarker. 

\begin{table}[t]
  \centering
\begin{tabular}{@{}c|ccc@{}}
\toprule
 \multirow{2}{*}{\textbf{Condition/Setup}}  & \textbf{$P_\beta$ $V_{Gi}$}  & \textbf{Power Usage}\\
&\textbf{$(\mu V^2 Hz^{-1})$} & \textbf{$(\mu A $ $cm^{-2} Hz)$}\\
 \midrule
 Healthy  & $348000$ &  $-$\\
 PD & $896000$  & $-$\\ 
 \midrule
 o-DBS \cite{carter2025invivotrainingdeepbrain} & $361000$  & $492$\\ 
 TD3 Deep-RL \cite{carter2025invivotrainingdeepbrain} & $336000$  & $341$\\ 
 \textbf{T3P MAB (ours)} & $189951$ & $216$\\ 
 \bottomrule
\end{tabular}
\begin{flushleft}
\vspace*{-0.4cm}
\end{flushleft}
 \caption{Comparison with existing approaches for DBS}    
      \label{tab:comparison_stimulation_power}
      \vspace*{-1cm}
\end{table}

In \cref{tab:comparison_stimulation_power} we compare the the energy consumption for stimulation in addition to patient outcome from an open-loop DBS (o-DBS), the TD3 deep-RL algorithm and our proposed T3P MAB. For the results, the initial exploration stage of the T3P MAB algorithm before the algorithm converges to an arm is not considered. Similarly, for the TD3 agent, we do not report values from the training phase but after it has been deployed for DBS. The comparison of patient outcomes is done in terms of $P_\beta$ of the $V_{Gi}$ signal from the BGT model. The power comparison is done as per the formula in \cref{eqn:power_consumption}. The results for the o-DBS setup and the TD3 Deep-RL are taken from \cite{carter2025invivotrainingdeepbrain}. For the o-DBS, a constant frequency of $130$ Hz and an amplitude of $2500$ $\mu A/cm^2$ is considered which is typical value used for most patients. The TD3 agent was trained on episodes of length $1000$ ms with a timestep of $100$ ms which allowed the updation of the stimulation parameters $10$ times during each episode. It was trained for a maximum duration of $5000$ timesteps or until the rewards converged. It uses a state representation consisting of the standard deviation of the synaptic conductance of the GPi neurons $S_{Gi}$, the Hjorth parameters (which include the activity, mobility and complexity) of the $S_{Gi}$ signal, $P_\beta$ of the GPi neurons and sample entropy, which is a measure of self-similarity of neuron readings of the STN. The reward function for the TD3 agent takes into account the energy consumption and the power spectral density (PSD) of the $S_{Gi}$ signal in the range $1-20$ Hz which is also a biomarker that is correlated with PD symptoms in a patient. We however we observe only the widely accepted $P_\beta$ from the GPi region for our reward function as discussed in \cref{sec:preliminaries}.

From \cref{tab:comparison_stimulation_power}, we can see that our MAB approach consumes less power while also being able to suppress the $P_\beta$ value better than the o-DBS and TD3 Deep-RL approaches. The RL agent was reported to use an average frequency and amplitude of $135$ Hz and $1690$ $\mu A/cm^2$ respectively. We found our approach to converge to the arm using a frequency of $155$ Hz and $1000$ $\mu A/cm^2$ in almost all cases. Additionally, it was observed that this combination always received more rewards on average than the other candidate parameter settings we consider for DBS. 

The TD3 agent relies on the PSD computation of the membrane potentials of the GPi neurons $V_{Gi}$ and $S_{Gi}$ signals sampled for a duration of $100$ ms. Through our experiments, we find that a duration of $100$ ms might not be long enough to obtain an accurate estimate of the PSD of a signal. Calculation of PSD based on signal lengths shorter than $500$ ms were found to be too unreliable for distinguishing between a more effective arm from a less effective one in the BGT model. This could also be an explanation of why the RL agent converged to a non-optimal frequency and amplitude setting as our approach beats the TD3 agent in both the aspects that were considered in its reward function --- power usage and alleviation of patient symptoms (quantified by the $P_\beta$ reading).  Note that we, in contrast, use a round length of $1000$ ms. Other possible reasons could be the unsuitability of the parameters used for state representation. A feature which is loosely correlated with patient outcome or associated with a lot of noise can stunt the ability of a deep-RL algorithm to learn the optimal policy. It is also worth noting that increasing the timestep length of an RL algorithm can lead to the requirement of a longer training time and/or more training data both of which can be difficult to achieve in practice.

An additional advantage of using MAB algorithms is the ability to quickly adapt to a case where the user is facing an unexpected side effect or discomfort with the specific parameter setting the algorithm has converged to. An example of this is shown in \cref{fig:user_intervention} where the user intervenes at round $75$. By the end of round $111$, it can be seen that the algorithm has converged to the second optimal arm, now that the optimal one has been pruned. This is however not straightforward with deep-RL techniques as the neural network component maps a state to an action and it would be difficult to choose an alternative action if the action returned is not acceptable for the patient.

\vspace*{-0.2cm}
\subsection{Implementation on Hardware}

\begin{table*}[t]
  \centering
\begin{tabular}{@{}ccc|ccccccc@{}}
\toprule

\multirow{2}{*}{\textbf{MCU}}  & \textbf{Clock} & \multirow{2}{*}{\textbf{Algorithm}} & \textbf{Round} & \textbf{Current} & \textbf{Voltage} & \textbf{Power} & \textbf{Energy} & \textbf{Time to} \\ 

& \textbf{Rate (MHz)} & & \textbf{Length (s)} & \textbf{Draw (mA)} & \textbf{(V)} & \textbf{Consumption (mW)} & \textbf{Usage (J)} & \textbf{Converge (s)} \\

\midrule

 \multirow{2}{*}{ESP32-S3} & \multirow{2}{*}{$240$} & T3P MAB & $1.2$ & $37.7$ & $5.13$ & $183.5$ & $5.4$ & $46.8$\\
  &  & T3P MAB* & $-$ & $31.6$ & $5.13$ & $162.4$ & $4.8$ & $-$ \\
\midrule

\multirow{2}{*}{ESP32-S3} & \multirow{2}{*}{$160$} & T3P MAB & $1.3$ & $32.0$ & $5.13$ & $164.1$ & $4.9$ & $50.7$\\
  &  & T3P MAB* & $-$ & $28.5$ & $5.13$ & $146.3$ & $4.3$ & $-$ \\
\midrule

 \multirow{2}{*}{ESP32-S3} & \multirow{2}{*}{$80$} & T3P MAB & $1.7$ & $27.6$ & $5.13$ & $141.5$ & $4.2$ & $66.3$\\
  &  & T3P MAB* & $-$ & $24.8$ & $5.13$ & $127.2$ & $3.8$ & $-$ \\
\midrule

 \multirow{2}{*}{ESP32-S3} & \multirow{2}{*}{$40$} & T3P MAB & $2.4$ & $17.8$ & $5.13$ & $91.9$ & $2.7$ & $93.6$\\
 &  & T3P MAB* & $-$ & $16.1$ & $5.13$ & $82.2$ & $2.4$ & $-$ \\
\midrule

 \multirow{2}{*}{ESP32-P4} & \multirow{2}{*}{$360, 360$} & T3P MAB & $1.1$ & $68.5$ & $5.12$ & $352.1$ & $10.4$ & $42.9$\\
 &  & T3P MAB* & $-$ & $67.3$ & $5.12$ & $344.7$ & $10.2$ & $-$\\
\midrule

 \multirow{2}{*}{ESP32-P4} & \multirow{2}{*}{$180, 180$} & T3P MAB & $1.2$ & $64.4$ & $5.12$ & $329.7$ & $9.8$ & $46.8$\\
 &  & T3P MAB* & $-$ & $63.2$ & $5.12$ & $323.4$ & $9.6$ & $-$\\
\midrule

\multirow{2}{*}{ESP32-P4} & \multirow{2}{*}{$90, 90$} & T3P MAB & $1.4$ & $58.9$ & $5.12$ & $302.1$ & $9.0$ & $54.6$\\
 &  & T3P MAB* & $-$ & $57.2$ & $5.12$ & $293.2$ & $8.7$ & $-$\\
\midrule

\multirow{2}{*}{ESP32-P4} & \multirow{2}{*}{$40, 40$} & T3P MAB & $1.9$ & $55.2$ & $5.12$ & $282.7$ & $8.4$ & $74.1$\\
& & T3P MAB* & $-$ & $54.4$ & $5.12$ & $278.5$ & $8.3$ & $-$\\
\bottomrule
\end{tabular}
\begin{flushleft}
* Indicates readings after the T3P MAB has converged to an arm 
\vspace*{-0.4cm}
\end{flushleft}
\caption{\centering Measurements with different clock frequencies on different MCUs}
\label{tab:hardware_comparison}
\vspace{-6mm}
\end{table*}

As described in \cref{experimental_setup}, we implement the T3P MAB algorithm on the ESP32-P4 and ESP32-S3 MCUs and measure energy metrics with Joulescope. For each round, $V_{Gi}$ signals are sampled at $100$ kHz, the highest rate permitted by the BGT model. The BGT model plays an arm with the stimulation parameters chosen by the algorithm while recording the $V_{Gi}$ signals for $1000$ ms. At the end of the round, $P_\beta$ from the $V_{Gi}$ signal is calculated and used to update the MAB algorithm parameters. 
\begin{wrapfigure}{r}{0.14\textwidth}
    {\centering
    \vspace*{-0.2cm}
    \hspace*{-0.3cm}\includegraphics[width=0.18\textwidth]{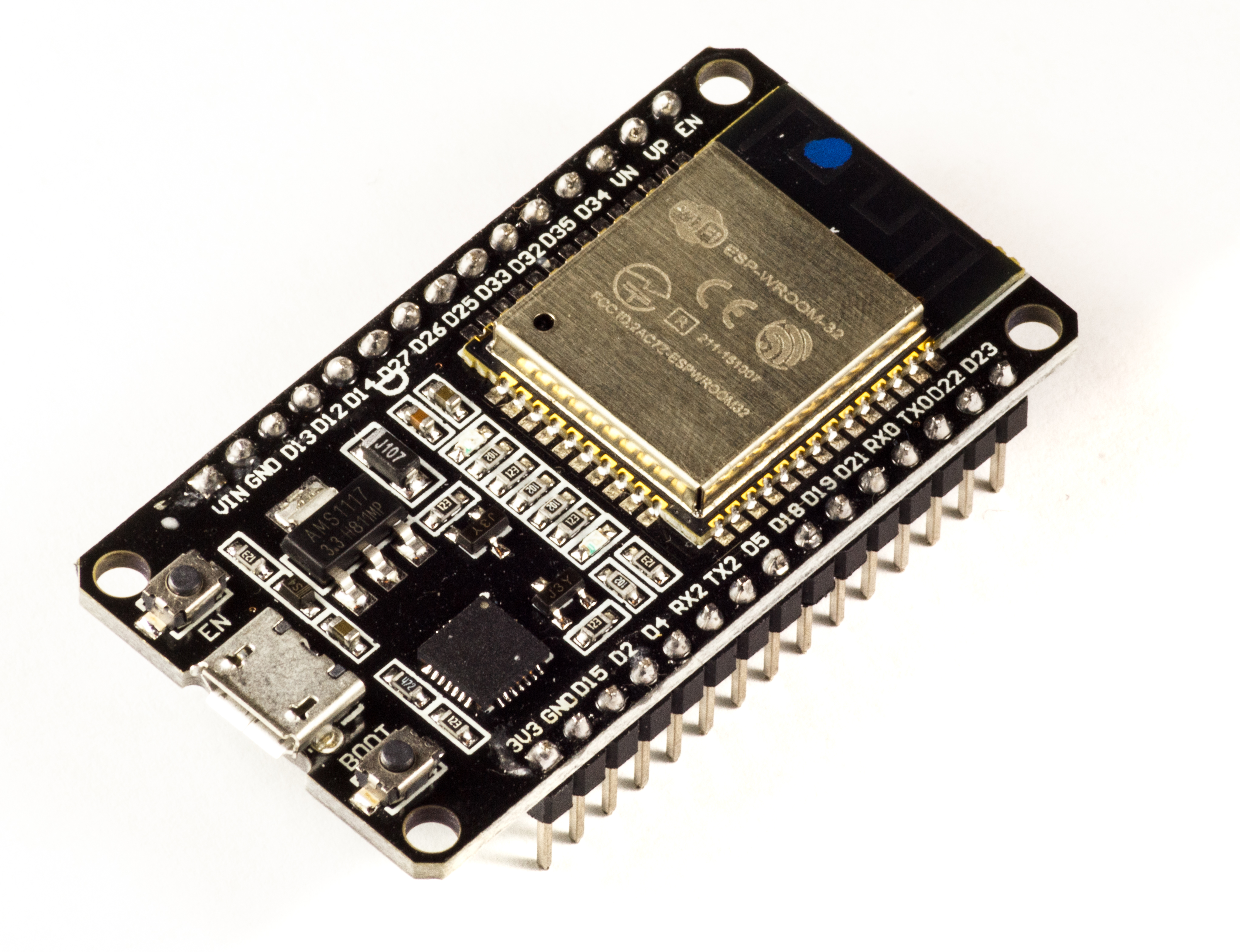}}
    \vspace*{-0.8cm}
    \caption{{Used MCU}}
    \vspace*{-0.2cm}
    \label{fig:ESP32-S3}
\end{wrapfigure}
We found $P_\beta$ calculation to be the most resource-intensive step of the algorithm. This is because finding $P_\beta$ after each round requires processing $100,000$ floating-point values. However, performing an FFT over all samples would require loading them into RAM simultaneously, which is not feasible. Therefore, we divide the signal into chunks and process them separately to compute the FFT and then $P_\beta$. Specifically, we used a Radix-2 Cooley--Tukey Decimation-in-Time (DIT) algorithm, which efficiently computes the discrete Fourier transform by recursively decomposing the input signal into even- and odd-indexed components and combining partial results using precomputed twiddle factors. This in-place iterative implementation minimizes computational overhead and memory usage, making it well-suited for deployment on hardware such as the ESP32 platform.

Now that we have discussed the details of our MCU implementation, we describe the statistics observed from the JoulescopeJS220 after hardware deployment. \cref{tab:hardware_comparison} lists observations from the MCUs at different clock speeds. For each case, we record time per episode, convergence time of the algorithm, current draw, voltage, power consumption, and time to converge to an arm while running the proposed algorithm. These energy measurements are based on a window of $30$ s. We experiment with $4$ different clock frequencies for both MCUs --- $240$, $160$, $80$, and $40$ with the ESP32-S3, and $360$, $180$, $90$, and $40$ with both cores in the ESP32-P4. The third low-power core on the ESP32-P4 was disabled for all experiments.

\begin{table*}[t]
  \centering
\begin{tabular}{@{}c|ccccccc@{}}
\toprule

\textbf{Attributes} & \textbf{Neurochip-2} & \textbf{PennBMBI} & \textbf{Toronto} & \textbf{Activa P+S} & \textbf{WAND} & \textbf{Summit RC+S} & \textbf{Ours}\\
\midrule

\textbf{Power} & $284-420$ mW & $290$ mW & $45$ mW & $500$ uW & $172$ mW & $2.5$ mW & $82.2$ mW\\
\textbf{Max. Current} & $5$ mA & $1$ mA & $3$ mA & $25.5$ mA & $5$ mA & $25.5$ mA & $5$ mA\\
\textbf{Sampling Rate} & $2/24$ kS/s & $21$ ks/s & $15$ kS/s & $422$ S/s & $1$ kS/s & $1$ kS/s & $10$ kS/s\\

\textbf{\makecell{Closed-loop Policy}} & \makecell{Detection/ \\ threshold-triggered} & \makecell{Detection/ \\ threshold-triggered} & Threshold & 2D SVM & Threshold & \makecell{8D SVM, \\ Threshold} & \makecell{RL \\ (MAB)}\\
\bottomrule
\end{tabular}
\begin{flushleft}
\vspace*{-0.4cm}
\end{flushleft}
\caption{\centering \footnotesize{Comparison with measurements form existing devices for Deep Brain Stimulation \cite{medtronic_summit_rc}.}}
\label{tab:existing_hardware}
\vspace{-10mm}
\end{table*}

It can be observed from the table that power consumption increases with clock frequency, as expected. The power draw, voltage, and energy readings are average values recorded over $30$ seconds. For every MCU and frequency, measurements from the T3P MAB agent are recorded. The first reading in every row records measurements when the T3P algorithm is running, and the second when the MAB algorithm has converged and is using a fixed stimulation parameter. A trade-off can be noticed here --- a setup that consumes less power is associated with longer convergence time. However, in all these cases, the algorithm takes less than two minutes to converge, which is clinically acceptable for patient use.

It is to be noted that it was not possible to train the TD3 deep-RL algorithm on the MCUs considered in our experiments. Although \cite{carter2025invivotrainingdeepbrain} discusses the possibility of \textit{in vivo} training in human brains, deep-RL algorithms take hours to train on traditional GPUs and would take even longer on MCUs given their limited computational capabilities. Hence, the goal of enabling personalized medicine by training the deep-RL agent \textit{in vivo} has not been realized in the work. 

Without the ability to train deep-RL agents inside the implant, they are traditionally trained offline and deployed only after their rewards have converged. This offline training process involves collecting patient data while using different stimulation parameters to create a dataset. This dataset is then used to train the RL agent before deploying it on the implant. This dataset is also prone to becoming outdated as the patient's brain changes over time (possibly with neurodegenerative progression) and might require training the RL agent all over again in the future. Our MAB approach, on the other hand, does not suffer from this limitation because it does not freeze parameters. Instead, it finds the optimal arm based on the new distribution of rewards in the current state of the brain. 

The MAB is expected to remain more energy efficient even when the TD3 deep-RL agent is deployed on the MCU after training as it avoids the requirement of inferencing a neural network every $100$ ms. Also, our T3P MAB algorithm relies only on the $P_\beta$ reading from the GPi neurons, in contrast to the $6$ features required by the TD3 agent for state representation, one of which requires an FFT to be performed, which clearly makes it more compute-intensive than our T3P MAB agent.

\subsection{Comparison with Existing Hardware}

Finally, we compare our most energy-efficient hardware implementation with DBS devices used in clinical trials. \Cref{tab:existing_hardware} lists Neurochip-2 \cite{neurochip2}, PennBMBI \cite{pennbmbi}, Toronto \cite{toronto1, toronto2, toronto3},
Activa P+S \cite{activa1,activa2,activa3,activa4}, WAND \cite{wand}, Summit RC+S \cite{medtronic_summit_rc}, and our T3P agent deployed on the ESP32-S3 at $40$ MHz, together with their power consumption, maximum DBS current, sampling rate, and closed-loop policy. Our setup consumes less power than three of the six other systems. Its maximum current is comparable to that of most devices, except the more recent Summit RC+S, which supports up to $25.5$ mA. Several devices with lower sampling rates also consume less power than our implementation, likely because processing more samples, especially for FFTs, is computationally intensive. Reducing the sampling rate for biomarker estimation may therefore be worthwhile, but is beyond the scope of this work.

Although some commercial DBS devices are highly power efficient, they still rely on threshold-based policies or SVMs. Such approaches typically require calibration using patient-specific data and offline parameter tuning, which can increase patient burden. Moreover, as discussed earlier, neural activity and biomarker characteristics may change over time, causing parameters selected during initial calibration to lose optimality and degrade long-term performance. In contrast, the RL-based approach in our work is a step toward precision medicine, enabling the implant to adapt to the current brain state and progression of PD rather than relying on outdated patient data. Most of these devices also use custom-designed MCUs, allowing careful hardware optimization to reduce current leakage and overall power consumption. Exploring such hardware co-design remains an important direction for future work.

\section{Concluding Remarks} \label{sec:conclusion}
In this work, we presented a resource-conscious adaptive framework for Deep Brain Stimulation (DBS) using lightweight reinforcement learning (RL) based on Multi-Armed Bandit (MAB) theory. We developed the Threshold-Triggered and Pruned (T3P) MAB algorithm and showed that clinically relevant DBS control can be achieved with lower computational complexity, faster convergence, and higher energy efficiency than deep reinforcement learning (deep-RL) approaches such as TD3. Our formulation unifies frequency and amplitude tuning within a single decision space, which remains largely unexplored in prior DBS literature. By jointly optimizing both parameters using neurophysiological biomarkers, the proposed method improves adaptability and patient safety while satisfying the hardware and energy constraints of implantable systems.

T3P improves upon conventional $\epsilon$-greedy methods by incorporating domain-specific characteristics, enabling rapid convergence while remaining responsive to changes in brain state. Our results show that it identifies optimal stimulation parameters faster than existing MAB approaches, outperforms deep-RL methods in suppressing the $P_\beta$ biomarker while using less power, and adapts robustly to user intervention when a patient experiences discomfort.

To our knowledge, this is the first hardware implementation of an MAB-based DBS controller with quantitative power and energy measurements on modern microcontrollers. In contrast, the TD3 deep-RL agent could not be feasibly trained or executed on such hardware because of its memory and computational requirements, highlighting the suitability of bandit-based methods for implantable devices. Overall, our T3P MAB algorithm delivers performance comparable to state-of-the-art deep-RL controllers at a fraction of their energy and computational cost, while enabling patient-specific adaptation without prior training in real-world clinical settings. Future work will explore reducing convergence time, implementing the full closed-loop setup in hardware, and experimenting with real patient data. We will also investigate the impact of sensing noise and possible reduction of biomarker-estimation overhead through lower sampling rates and hardware--algorithm co-design to further improve implant energy efficiency. \\

\vspace*{-0.2cm}
\noindent 
{\bf Acknowledgments:} We thank all the reviewers for their constructive comments. This work was partially supported by the German Research Foundation (DFG) project number 571618200 and by a Dieter Schwarz Courageous Research Grant on Sustainable Cyber-Physical Systems.

\bibliographystyle{IEEEtran}
\bibliography{refs}

\end{document}